\documentclass[runningheads]{llncs}
\usepackage{color}
\usepackage{graphicx}
\usepackage{hyperref}
\usepackage{amsmath}
\usepackage{amssymb}

\usepackage{array}
\usepackage{booktabs}
\usepackage{multirow}
\usepackage{makecell}
\usepackage{ulem}
\usepackage[misc]{ifsym}

\usepackage{bbm}

\begin{document}

\title{
LGS:
A Light-weight 4D Gaussian Splatting for Efficient Surgical Scene Reconstruction
}

%
%
%
\authorrunning{Liu et al.}
\titlerunning{Light-weight 4D Gaussian Splatting for Surgical Scene Reconstruction}
\author{
{Hengyu Liu\thanks{Equal contribution},  Yifan Liu\textsuperscript{$\star$}, Chenxin Li\textsuperscript{$\star$}, Wuyang Li, and Yixuan Yuan$^{(\textrm{\Letter})}$}}
\institute{The Chinese University of Hong Kong\\
\email{hy.liu321@gmail.com, yxyuan@ee.cuhk.edu.hk}
}

%
%
\maketitle              
\begin{abstract}
The advent of 3D Gaussian Splatting (3D-GS) techniques and their dynamic scene modeling variants, 4D-GS, offers promising prospects for real-time rendering of dynamic surgical scenarios. However, the prerequisite for modeling dynamic scenes by a large number of Gaussian units, the high-dimensional Gaussian attributes and the high-resolution deformation fields, all lead to serve storage issues that hinder real-time rendering in resource-limited surgical equipment. To surmount these limitations, we introduce a Lightweight 4D Gaussian Splatting framework (LGS) that can liberate the efficiency bottlenecks of both rendering and storage for dynamic endoscopic reconstruction.
Specifically, to minimize the redundancy of Gaussian quantities, we propose Deformation-Aware Pruning by gauging the impact of each Gaussian on deformation. 
Concurrently, to reduce the redundancy of Gaussian attributes, we simplify the representation of textures and lighting in non-crucial areas by pruning the dimensions of Gaussian attributes. 
We further resolve the feature field redundancy caused by the high resolution of 4D neural spatiotemporal encoder for modeling dynamic scenes via a 4D feature field condensation. 
Experiments on public benchmarks demonstrate efficacy of LGS in terms of a compression rate exceeding 9$\times$ while maintaining the pleasing visual quality and real-time rendering efficiency.
LGS confirms a substantial step towards its application in robotic surgical services. Project page: \url{https://lgs-endo.github.io/}.
\keywords{3D Reconstruction \and Gaussian Splatting \and Robotic Surgery.}
\end{abstract}
\section{Introduction}

 Reconstructing dynamic 3D scenes from endoscopic videos holds paramount significance in minimally invasive surgeries~\cite{wang2022neural,zha2023endosurf,yang2023mrm}, as it enhances comprehension of the spatial environment surrounding the surgical site, thereby enabling surgeons to conduct more precise and efficient operations~\cite{mahmoud2017orbslam,li2024endora,he2023h,liu2023efficientvit}. Concomitantly, this technology fosters a myriad of subsequent applications, encompassing Virtual Reality (VR) surgeries, medical pedagogy, and the automation of robotic surgeries~\cite{lu2021super,penza2017envisors,tang2018augmented,ali2024assessing}. Subsequent advancements, particularly real-time rendering, have emerged as a cutting-edge methodology conducive to deployment on robotic apparatus, thereby facilitating the progression of robotic surgical automation~\cite{yu2022tnn,liu2023grab,li2022scan}, medical imaging system~\cite{sun2022few,li2021unsupervised,he2023h} and aided diagnosis procedure~\cite{li2021htd,liu2021consolidated,yu2019dense}.


Previous 3D reconstruction methods encompass various approaches, including those employing depth estimation~\cite{brandao2021hapnet,luo2022unsupervised,liu2024stereo}, SLAM-based methods~\cite{song2017dynamic,zhou2021emdq}, and techniques leveraging a sparse warp field \cite{li2020super,gao2019surfelwarp}. 
With the rising of neural rendering, Neural Radiance Fields (NeRF)~\cite{mildenhall2021nerf,barron2021mip,li2023steganerf} is introduced to reconstruct the surgical scenes\cite{wang2022neural,yang2023neural,li2023novel,zha2023endosurf}, unlocking the limited reconstruction quality of dynamic scenes compared with conventional methods. 
EndoNeRF \cite{wang2022neural} and its following works~\cite{yang2023neural,zha2023endosurf} utilize dynamic neural radiance fields to model deformed surgical scenes and achieve satisfied rendering quality. Recently, 3D Gaussian Splatting (3D-GS) \cite{kerbl20233d}, which utilizes an explicit 3D Gaussian representation with specific attributes to model scenes and a differentiated splatting-based rendering technique,
has exhibited pleasing reconstruction efficiency from a series of static captures. 
Further advances of 4D-GS~\cite{wu20234dgaussians} break this \textit{static} limitation by additionally  introduce an elaborated high-resolution spatiotemporal feature field to model the time-varying deformation~\cite{wu20234dgaussians} and the relevant progress has been extended to clinical scenarios like reconstructing deformable tissues in 
dynamic endoscopic scenes in real-time efficiency~\cite{chen2024endogaussians,huang2024endo,liu2024endogaussian,zhuendogs,zhang2021generator}.

Nevertheless, owing to using explicit point-based representations and intricate high-dimensional spatial-temporal fields, the storage burden of 3D representation is significantly enlarged, thereby limiting the practical deployment on resource-limited surgical devices and robots~\cite{ahmed2023pre,sherif2023remote,li2022knowledge,li2022sigma,yu2022break}.  
In particular, the memory burden by 4D-GS primarily lies in the following folds.
(\textbf{a}) {Redundant Number of Gaussians}: 
Gaussian densification~\cite{kerbl20233d} increases the number of Gaussian for the accurate reconstruction of granular details, which results in a significant memory cost~\cite{fan2023lightgaussian}. While compared with nature scenes, surgical scenarios require less Gaussians to model the relatively simple environment.
(\textbf{b}) {Redundant Dimension of Gaussian Attributes}:
Natural scenes require high-dimensional attributes to represent the rich textures and varying illumination. While the intricate environment of surgical scenes such as repetitive textures and view-dependent lighting~\cite{batlle2023lightneus,rodriguez2022tracking} can be represented with fewer parameters, which causes the redundant dimension of Gaussian Attributes. 
(\textbf{c}) {High Resolution of Spatial-temporal Fields}: A lower-resolution feature field \cite{peng2021low} is sufficient to represent deformation, while the high resolution used in spatial-temporal feature fields is redundant to model the dynamics and encode the endoscopic scenario information.

To address the challenges of practical deployment of reconstruction models posed by the aforementioned prohibitively high memory burden.
we propose a holistic Lightweight 4D Gaussian Splatting (LGS) framework that allows for achieving satisfactory endoscopic reconstruction with both efficient rendering and storing~\cite{hinton2015distilling,li2024u}. 
In particular, to alleviate the \textbf{Quantity} burden of Gaussian representation, we present a Deformation-Aware Pruning (DAP) which identifies the informative Gaussians by a carefully designed deformation score
 and reduces the redundant Gaussians based on a deformation perception. To address the  \textbf{High-dimension} burden of Gaussian attributes, we propose a Gaussian-Attribute Pruning (GAP), which elucidates the uninformative visual patterns modeled by high-frequency components (non-critical textures) in Spherical Harmonic attributes, thereby steering the representation capabilities towards low-frequency attribute characterization.
 To mitigate the  \textbf{High-resolution} burden of spatial-temporal deformable fields, we present Feature Field Condensation (FFC) which compactly represents the high-resolution spatial-temporal feature field and performs an adaptive quantized projection of spatial-temporal coordinates. 
 Experimental results show that LGS can achieve higher storage efficiency with an over $9\times$ compression rate, whilst maintaining pleasing reconstruction quality and rendering speed, confirming the efficacy of our framework for practical deployment in clinical robotic surgical equipment.

\begin{figure}[t]
    \centering 
    \includegraphics[width=12cm]{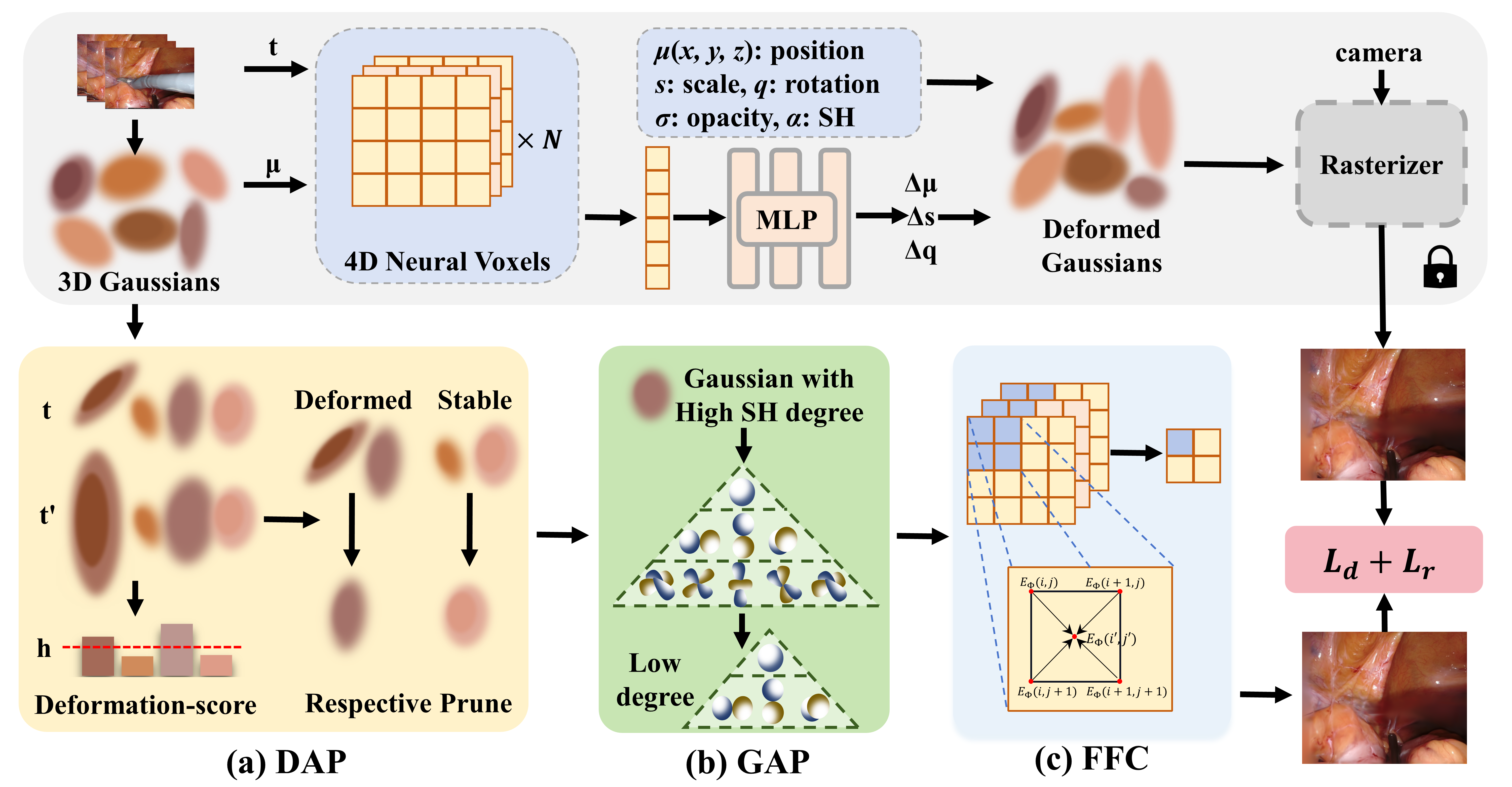}
    \caption{\textbf{LGS Overview} consists of (a) Deformation-Aware Pruning, (b) Gaussian-Attribute Pruning, (c) Feature Field Condensation, and distillation for optimization.}
    \label{fig:model}
\end{figure}

\section{Method}

The overview of LGS is shown in Fig.~\ref{fig:model}. In this section, we first introduce the representation of 3D-GS \cite{kerbl20233d} and its dynamic variant \cite{wu20234dgaussians} in Sec.~\ref{sec_preliminaries}. Then we introduce the DAP and GAP respectively in Sec.~\ref{sec_Deformation-aware Pruning} and Sec .~\ref {sec_SH Pruning}. In Sec.~\ref{sec_3D Feature Field Condensation} we conduct FFC and finally, we describe the optimization of LGS in Sec.~\ref{sec_optimization}.

\subsection{Preliminaries}\label{sec_preliminaries}

\noindent{\textbf{3D-GS}}
As introduced in \cite{kerbl20233d}, a set of dense Gaussians is utilized to represent the 3D data and achieve real-time rendering of dynamic scenes. Each Gaussian is defined by its center $\boldsymbol{\mu} \in \mathbb{R}^3$ and the covariance matrix $\boldsymbol{\Sigma}\in \mathbb{R}^{3\times 3}$, which can be decomposed into a scaling factor $\boldsymbol{s} \in \mathbb{R}^3$ and a rotation quaternion $\boldsymbol{q} \in \mathbb{R}^4$ for differentiable optimization. Colors and view-dependent appearance can be represented by opacity $\sigma \in \mathbb{R}$ and SH coefficients $\boldsymbol{\alpha} \in \mathbb{R}^C$ ($C$ devotes the number of SH). After differential splatting \cite{yifan2019differentiable} and adaptive density control \cite{kerbl20233d}, these Gaussians are optimized to achieve real-time rendering for dynamic scenes,

\begin{equation}
\label{eq1}
\text{G}(\boldsymbol{x})=e^{-\frac{1}{2}(\boldsymbol{x - \mu})^T\boldsymbol{\Sigma}^{-1} (\boldsymbol{x - \mu})}
\end{equation}

\noindent{\textbf{Dynamic Scene Rendering}}
As the extension of 3D-GS, \cite{wu20234dgaussians} utilizes a deformation module which involves a spatial-temporal feature field $E$ and a tiny MLP $F$ to calculate deformation. Given a 4D input consisting of the Gaussian center $\boldsymbol{\mu}= (x, y, z)$ and  query time $t$, the feature field $E$ retrieves the latent feature $f$ of the input: $\boldsymbol{f} = E(\boldsymbol{\mu}, t)$. Subsequently, the tiny MLP $F$ calculates the deformation in position, rotation, and scaling: $\Delta{\boldsymbol{\mu}}, \Delta{\boldsymbol{q}}, \Delta{\boldsymbol{s}} = F(\boldsymbol{f})$. Then the $i$-th Gaussian can be represented as $\boldsymbol{\Theta_i}=\{\boldsymbol{\mu_i}', \boldsymbol{q_i}', \mathbf{s_i}', \sigma_i, \boldsymbol{\alpha_i}\}$, where $\boldsymbol{\mu_i}', \boldsymbol{q_i}', \boldsymbol{s_i}' = \boldsymbol{\mu_i} + \Delta{\boldsymbol{\mu_i}}, \boldsymbol{q_i} + \Delta{\boldsymbol{q_i}}, \boldsymbol{s_i} + \Delta{\boldsymbol{s_i}}$.

\subsection{Deformation-Aware Pruning}
\label{sec_Deformation-aware Pruning}



To distinguish Gaussians that are informative for modeling deformable dynamics, we propose a deformation score to identify the impact of each Gaussian on deformation enabling the follow-up importance-based pruning can be conducted.

\noindent{\textbf{Deformation Score}}\label{Deformation-score Calculation}
Inspired by the color rendering in 3D-GS \cite{kerbl20233d} and the global importance score in \cite{fan2023lightgaussian}, the deformation score of each Gaussian is associated with its contribution to all pixels and its deformation in volume across all timestamps. We use the criterion $\mathbbm{1}({G}(\boldsymbol{X}_i), \boldsymbol{p}_{k}, t)$ to reflect whether the $i$-th Gaussian is contributed to the pixel $k$ at timestamp $t$. The volume deformation of the $i$-th Gaussian $\Delta{V(\boldsymbol{s_i})}$ can be calculated with scaling factor $\boldsymbol{s_i}$ and deformation $\Delta\boldsymbol{s_i}$.
Consequently, the deformation score is obtained by,
\begin{equation}
\label{deformation-score-cal}
d_i = \sum_{t}^{T}\sum_{k}^{HW}{\mathbbm{1}({G}(\boldsymbol{X}_i), \boldsymbol{p}_{k}, t) \cdot \Delta{V(\boldsymbol{s_i})}},
\Delta{V(\boldsymbol{s_i})} = \sum_t^{T} \Vert V(\boldsymbol{s_i}) -  V(\boldsymbol{s_i} + \boldsymbol{\Delta{s_i}})\Vert_1
\end{equation}
where $T$, $H$, and $W$ denotes the timestamps, height and weight of the image, $V(\boldsymbol{s}) = 4\pi \boldsymbol{s}_1 \boldsymbol{s}_2 \boldsymbol{s}_3/3$ devotes the volume of Gaussian with $\boldsymbol{s}$.

\noindent{\textbf{Deformation-Aware Pruning}}\label{}
Based on the deformation score and a threshold $h$, Gaussians are classified into two categories: stable $SG=\{i | d_i \le h \}$ and deformed $DG=\{i | d_i > h\}$. Then we conduct DAP respectively on $DG$ and $SG$: as shown in Eq.\ref{import-score-cal}, important score is computed for $SG$ based on opacity and normalized volume, and for $DG$ based on original volume and its deformation,
\begin{equation}
\label{import-score-cal}
IS_i = \left\{
	\begin{aligned}
	 \sum_{t}^{T}\sum_{k}^{HW}{\mathbbm{1}({G}(\boldsymbol{X}_i), \boldsymbol{p}_{k}, t) \cdot \sigma_i \cdot V_{norm}(\boldsymbol{s_i}) }, \quad i \in SG \\
	 \sum_{t}^{T}\sum_{k}^{HW}{\mathbbm{1}({G}(\boldsymbol{X}_i), \boldsymbol{p}_{k}, t) \cdot \Delta{V(\boldsymbol{s})} \cdot V_{norm}(\boldsymbol{s_i}) },  \quad i \in DG \\
	\end{aligned}
	\right.
\end{equation}
where $V_{norm}(\boldsymbol{s}) =  (V(\boldsymbol{s})/V_{max90})^{\beta}$ reflects the normalized volume with a norm factor $\beta$ and $V_{max90}$ reflects the $90\%$ largest volume of all sorted Gaussians. Then we prune the Gaussians with lower important scores in each of the two classes accordingly, which can remove unimportant Gaussians while retaining those crucial for deformation.


%


\subsection{Gaussian-Attribute Pruning}
\label{sec_SH Pruning}
Gaussian Attributes are redundant to represent the intricate environment of surgical scenes caused by repetitive textures and view-dependent lighting. SH coefficients contain 48 floating-point values and represent over $80\%$ of all attributes for each Gaussian \cite{fan2023lightgaussian}, which is far more than other attributes that cannot be pruned, such as position, rotation, scaling factor, and opacity. Based on this, GAP reduces the higher degree SH coefficients used to model the view-dependent color and scene reflection.  To enhance model memory efficiency and suitability for surgical scenes, we use a threshold to represent the pruned degree of SH coefficients~\cite{fan2023lightgaussian,xu2022afsc,pan2023learning}. As described in Eq.~\ref{SH-pruning-cal}, GAP adjusts the SH degree from high to low, which effectively reduces the redundancy of each Gaussian's attributes.
\begin{equation}
    \label{SH-pruning-cal}
    \alpha_{ic} = \alpha_{ic} * \mathbbm{1}(c \le (h_{sh} + 1)^2 + N_{RGB}), c \in C
\end{equation}
where $h_{sh}$ is the threshold for low SH degree, and $N_{RGB}$ represents the number of SH coefficients to depict RGB colors, which is normally set to $3$. To take full advantage of the information contained in the pruned attributes for modeling the surface of the object, we utilize distillation to transfer knowledge from SH coefficients at higher degree, which will be detailed in Sec.~\ref{sec_optimization}.

\subsection{Feature Field Condensation}
\label{sec_3D Feature Field Condensation}
The deformation module including a spatial-temporal feature field $E$ and a tiny MLP $F$ is used to learn the representation of 4D data and the deformation at different timestamps. 
Despite the existence of methods such as Gaussian pruning and vector quantization \cite{fan2023lightgaussian,lee2023compact} that aim to reduce the size of GS-based models, they are primarily designed for static scenes and are unable to address memory issues. To represent the position of the Gaussian and the query time, 
a spatial-temporal feature field with higher resolution is used for 4D encoding. Though this approach allows the model to capture more details of deformation, it also requires a significant amount of storage space, accounting for over $80\%$ of the required storage, which hinders model deployment on robotic surgical devices. 

Motivated by these observations, we propose the FFC to make our model more memory-efficient and easier to deploy on robotic surgical devices. Considering that nearby 3D Gaussians always share similar spatial and temporal information which means adjacent values on the same voxel plane should also be similar. To reduce the size of the voxel plane and preserve every 4D feature as much as possible, we conduct 3D adaptive pooling on each 4D voxel plane, which is described in Eq. \ref{hxplane-cal}. The tiny MLP $F$ remains to preserve the ability of deformation calculation,
\begin{equation}
\label{hxplane-cal}
E_{\Phi}'(i, j) =  {1\over{r_{\Phi_1} r_{\Phi_2}}}\sum_{i'=i\cdot r_{\Phi_1}}^{(i + 1)\cdot r_{\Phi_1}}\sum_{j'=j\cdot r_{\Phi_2}}^{(j + 1)\cdot r_{\Phi_2}}{E_{\Phi}(i',j')}
\end{equation}
where $\Phi=(\Phi_1, \Phi_2) \in \{(x,y),(x,z),(y,z),(x,t),(y,t),(z,t)\}$ devotes the sub-planes of the spatial-temporal feature field, and $r_{\Phi_i}$represents the compression rate of the feature field on the $\Phi_i$ axis.
\subsection{Optimization}
\label{sec_optimization}
To achieve the balance between efficient memory and high-quality rendering performance, we use knowledge distillation \cite{hinton2015distilling,li2022knowledge} during optimization. We treat the uncompressed and well-trained model as teacher model, and the model processed by DAP, GAP, and FFC as the student model. 
We minimize the loss $L = L_{d} + L_{r}$ to better transfer knowledge from the trained teacher model to the memory-efficient student model and achieve a memory-efficient and high-quality model~\cite{hinton2015distilling,liang2021unsupervised}. $L_{d}$ is the distillation loss between the rendered images of teacher model and student model, and $L_{r}$ is the rendering loss between the rendered image of student model and the ground truth,
\begin{equation}
\label{distill_loss-cal}
L_{d} = \frac{1}{T}\sum_{t}^{T} \Vert \boldsymbol{\hat{I}_{tch}}(t) - {\boldsymbol{\hat{I}_{stu}}(t)}\Vert_2, \quad L_{r} = \frac{1}{T}\sum_{t}^{T} \Vert \boldsymbol{{I}_{gt}}(t) - {\boldsymbol{\hat{I}_{stu}}(t)}\Vert_2
\end{equation}
where $T$ is the number of training timestamps,  $\boldsymbol{\hat{I}_{tch}}(t)$,  $\boldsymbol{\hat{I}_{stu}}(t)$ and $\boldsymbol{{I}_{gt}}(t)$ are the rendered image of teacher model and student model, and the ground truth image at timestamp $t$.


\section{Experiments}
\subsection{Experimental Settings}

\noindent{\textbf{Datasets}}
We conduct experiments on two widely used datasets: 
ENDONERF \cite{wang2022neural} and SCARED \cite{allan2021stereo}. ENDONERF \cite{wang2022neural} consists of two public cases of in-house DaVinci robotic prostatectomy data, each depicting a single-view scene with non-rigid deformation and tool occlusion. Following \cite{zha2023endosurf}, we use 5 keyframes of  SCARED \cite{allan2021stereo} captured by a da Vinci Xi surgical robot. To align with prior work \cite{liu2024endogaussian,zha2023endosurf}, we divided each keyframe into 7:1 training and testing sets.


\noindent{\textbf{Compared Methods}}
We compare with recent reconstruction methods of dynamic surgical scenes, including NeRF-based methods: EndoNeRF\cite{wang2022neural}, EndoSurf \cite{zha2023endosurf}, LerPlane \cite{yang2023neural}, and GS-based methods: EndoGaussian \cite{liu2024endogaussian}, EndoGS\cite{zhuendogs}. We use PSNR, SSIM, and LPIPS as metrics for rendering quality. Following \cite{fan2023lightgaussian}, we use the storage size of the model and frames-per-second (FPS) during inference as metrics for memory efficiency and rendering speed respectively.


\begin{table}[t]
  \centering
  \caption{Experimental results on ENDONERF \cite{wang2022neural} and SCARED \cite{allan2021stereo}}
    
    \begin{tabular}{ccccccc}
    \toprule
    Dataset & Method & Size(↓) & FPS(↑) & SSIM(↑) & PSNR(↑) & LPIPS(↓) \\
    \midrule
    \multirow{7}[2]{*}{ENDONERF\cite{wang2022neural}} & EndoNeRF\cite{wang2022neural} & 13.00MB    & 0.035  & 0.933  & 36.06  & 0.089  \\
          & EndoSurf\cite{zha2023endosurf} & 20.00MB    & 0.040  & 0.954  & 36.53  & 0.074  \\
          & LerPlane-9k\cite{yang2023neural} & 274.0MB   & 0.911  & 0.926  & 34.99  & 0.080  \\
          & LerPlane-32k\cite{yang2023neural} & 274.0MB   & 0.872  & 0.950  & 37.38  & 0.047  \\
          & EndoGS\cite{zhuendogs} & 322.7MB & 91.75  & 0.963  & 37.29  & 0.045  \\
          & EndoGaussian\cite{liu2024endogaussian} & 334.5MB & 166.5  & 0.960  & 37.78  & 0.053  \\
          & LGS (Ours) & 25.00MB    & 188.3  & 0.955  & 37.48  & 0.068  \\
    \midrule
    \multirow{4}[2]{*}{SCARED\cite{allan2021stereo}} & EndoNeRF\cite{wang2022neural} & 6.900MB   & 0.016  & 0.768  & 24.35  & 0.397  \\
          & EndoSurf\cite{zha2023endosurf} & 14.00MB    & 0.009  & 0.802  & 25.02  & 0.356  \\
          & EndoGaussian\cite{liu2024endogaussian} & 184.0MB   & 170.56  & 0.825  & 26.89  & 0.272  \\
          & LGS (Ours) & 20.40MB  & 194.66  & 0.826  & 27.05  & 0.297  \\
    \bottomrule
    \end{tabular}%
  \label{tab-main-results}%
\end{table}%

\begin{table}[t]
  \centering
  \caption{Ablation Study on ENDONERF \cite{wang2022neural} for each component of LGS}
    \begin{tabular}{c|cccc|ccc}
    \toprule
    Model & Overall Size↓ & GS Size↓ & Deform Size↓ & FPS↑ & SSIM↑ & PSNR↑ & LPIPS↓ \\
    \midrule
    w/o DAP & 56.70MB & 33MB  & 23.70MB & 74.44 & 0.963  & 38.36  & 0.051  \\
    w/o GAP & 26.30MB & 5.0MB & 21.30MB & 187.9 & 0.950 & 37.08  & 0.096  \\
    w/o FFC & 329.6MB & 3.3MB & 326.3MB & 177.9 & 0.945 & 35.82  & 0.082  \\
    Full model (Ours) & \textbf{24.50MB} & \textbf{3.3MB} & \textbf{21.20MB} &\textbf{188.5} & 0.957  & 38.08  & 0.079  \\
    \bottomrule
    \end{tabular}%
  \label{tab-ablation}%
\end{table}%

\noindent{\textbf{Implementation Details}} We use our checkpoints with the training details described in EndoGaussian \cite{liu2024endogaussian}. For DAP, we set the $h$ to $0.5$ and $\beta$ to $0.1$. For GAP, we set $h_{sh}$ to 2. For FFC, we set the resolution of the feature field to $[16,16,16,25]$. We implement our framework with Pytorch and use the differential rasterization \cite{fan2023lightgaussian} as the render engine.

\subsection{Experimental Results}

\begin{figure}[t]
    \centering 
    \includegraphics[width=12cm]{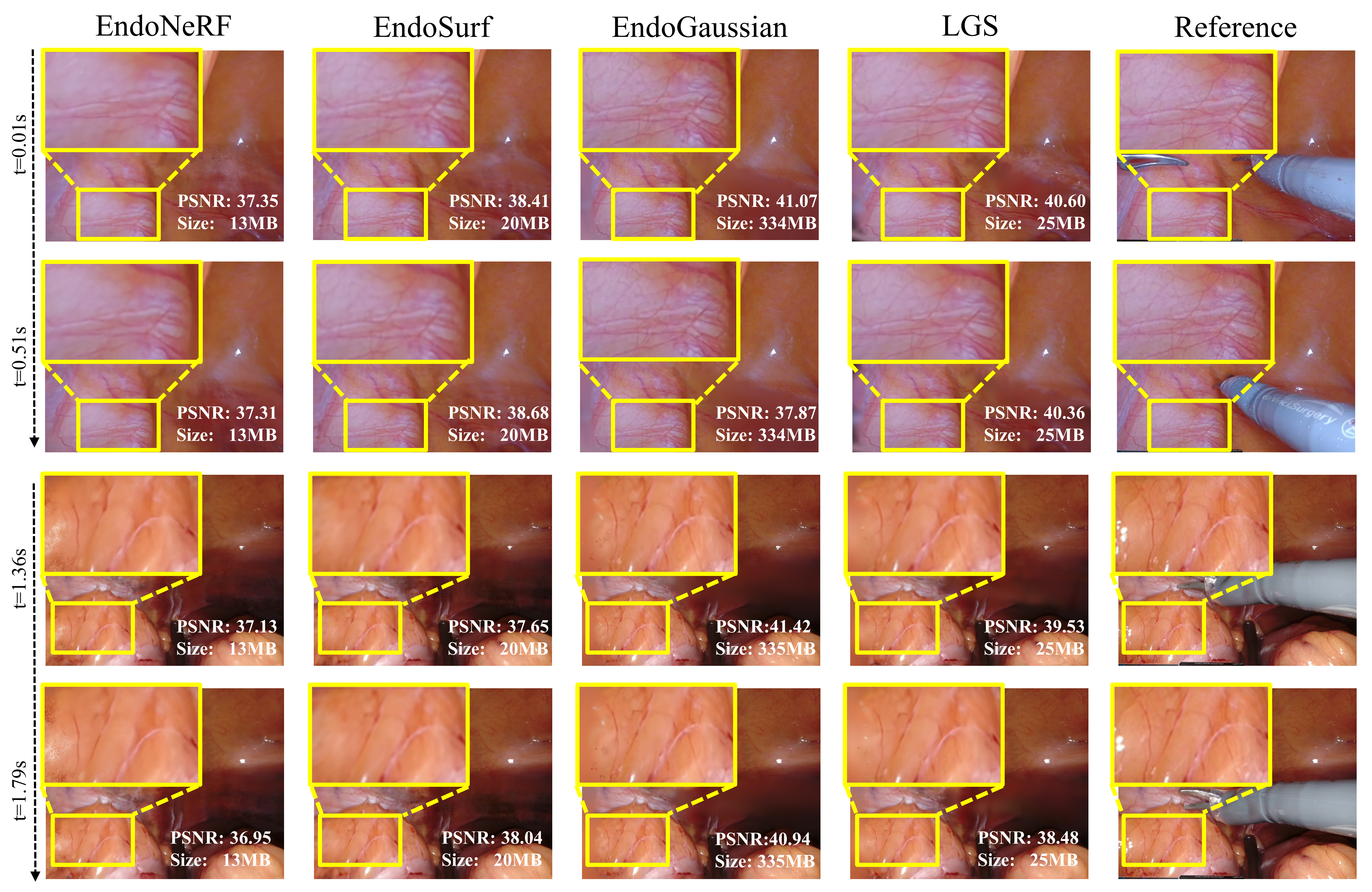}
    \caption{Rendered images of previous methods and ours: PSNR reflects the quality of the shown image and Size reflects the used memory to store model.}
    \label{fig:result}
\end{figure}

The experimental results on ENDONERF \cite{wang2022neural} and SCARED \cite{allan2021stereo} are presented in Table \ref{tab-main-results}. It can be observed that the model size of LGS achieved a compression rate of $15\times$ and $9\times$ compared to EndoGaussian \cite{liu2024endogaussian} on ENDONERF and SCARED, respectively. 
Moreover, LGS exhibits a model size on the same order of magnitude as EndoNeRF \cite{wang2022neural} and EndoSurf \cite{zha2023endosurf}, inferring that LGS is as memory-efficient as NeRF-based methods. 
The rendering quality difference between LGS and GS-based methods can be ignored with $0.3$ lower in PSNR and $0.05$ lower in SSIM on ENDONERF, while LGS outperforms EndoGaussian on SCARED in SSIM and PSNR. Moreover, the visual results in Fig.~\ref{fig:result} show that LGS can render the details in surgical scenes as well as the GS-based method. LGS has a higher rendering speed than EndoGaussian and EndoGS, $21.8$ and $87.9$ higher on FPS respectively. However, there is a clear gap between GS-based methods and NeRF-based methods in terms of rendering speed and quality. Therefore, LGS can achieve high-quality and real-time rendering for dynamic surgical scenes with memory efficiency similar to NeRF-based methods.


\subsection{Ablation Studies}
To prove the effectiveness of each component of LGS, we conduct ablation experiments on the ENDONERF \cite{wang2022neural} dataset. and experimental results are presented in Tab.~\ref{tab-ablation}. We can observe that DAP prunes the least unimportant Gaussians, GAP reduces the number of parameters of each Gaussian by $40\%$, and FFC is also significant for memory efficiency, as it reduces the resolution of spatial-temporal feature
field. The use of GAP and FFC can also improve the performance of the model, which respectively solve the repetitive textures mentioned in Sec.~\ref{sec_SH Pruning} and utilize the spatial-temporal similarity of nearby Gaussians mentioned in Sec.~\ref{sec_3D Feature Field Condensation}. Moreover, Tab. \ref{tab-ablation} suggests that the three proposed methods all contribute positively to the enhancement of FPS. Overall, the three parts of LGS are all important to achieve high-quality and real-time rendering of dynamic surgical scenes with memory efficiency.

\section{Conclusion}

This paper presents Light-weight 4D Gaussian Splatting (LGS), a breakthrough method for endoscopic reconstruction that addresses the challenges of rendering efficiency and memory efficiency. Specifically, we employ Deformation-Aware Pruning to minimize the redundancy of unnecessary Gaussian quantities and conduct Gaussian-Attribute Pruning to prune the dimensions of Gaussian attributes which simplifies the representation of textures and lighting in non-crucial areas. Additionally, we use Feature Field Condensation to tackle high-resolution redundancy in the spatial-temporal feature field for dynamic scenes. Our experiments on public datasets demonstrate that LGS achieves a compression rate exceeding 9 $\times$ while maintaining high-quality and real-time rendering. With LGS, we take a significant step towards the practical deployment of 4D Gaussian Splatting in robotic surgical services.

\subsubsection{\ackname}We thank Zhiwen Fan from The University of Texas at Austin for valuable discussions on extending neural rendering techniques to the medical image domain, which greatly contributed to this work. This work was supported by Hong Kong Innovation and Technology Commission Innovation and Technology Fund ITS/229/22 and Research Grants Council (RGC) General Research Fund 14204321, 11211221.

\bibliographystyle{splncs04}
\bibliography{reference}

\begin{thebibliography}{10}
\providecommand{\url}[1]{\texttt{#1}}
\providecommand{\urlprefix}{URL }
\providecommand{\doi}[1]{https://doi.org/#1}

\bibitem{ahmed2023pre}
Ahmed, M.S., Giordano, S.: Pre-trained lightweight deep learning models for surgical instrument detection: Performance evaluation for edge inference. In: GLOBECOM 2023-2023 IEEE Global Communications Conference. pp. 3873--3878. IEEE (2023)

\bibitem{ali2024assessing}
Ali, S., Ghatwary, N., Jha, D., Isik-Polat, E., Polat, G., Yang, C., Li, W., Galdran, A., Ballester, M.{\'A}.G., Thambawita, V., et~al.: Assessing generalisability of deep learning-based polyp detection and segmentation methods through a computer vision challenge. Scientific Reports  \textbf{14}(1), ~2032 (2024)

\bibitem{allan2021stereo}
Allan, M., Mcleod, J., Wang, C., Rosenthal, J.C., Hu, Z., Gard, N., Eisert, P., Fu, K.X., Zeffiro, T., Xia, W., et~al.: Stereo correspondence and reconstruction of endoscopic data challenge. arXiv preprint arXiv:2101.01133  (2021)

\bibitem{barron2021mip}
Barron, J.T., Mildenhall, B., Tancik, M., Hedman, P., Martin-Brualla, R., Srinivasan, P.P.: Mip-nerf: A multiscale representation for anti-aliasing neural radiance fields. In: Proceedings of the IEEE/CVF International Conference on Computer Vision. pp. 5855--5864 (2021)

\bibitem{batlle2023lightneus}
Batlle, V.M., Montiel, J.M., Fua, P., Tard{\'o}s, J.D.: Lightneus: Neural surface reconstruction in endoscopy using illumination decline. In: International Conference on Medical Image Computing and Computer-Assisted Intervention. pp. 502--512. Springer (2023)

\bibitem{brandao2021hapnet}
Brandao, P., Psychogyios, D., Mazomenos, E., Stoyanov, D., Janatka, M.: Hapnet: hierarchically aggregated pyramid network for real-time stereo matching. Computer Methods in Biomechanics and Biomedical Engineering: Imaging \& Visualization  \textbf{9}(3),  219--224 (2021)

\bibitem{chen2024endogaussians}
Chen, Y., Wang, H.: Endogaussians: Single view dynamic gaussian splatting for deformable endoscopic tissues reconstruction. arXiv preprint arXiv:2401.13352  (2024)

\bibitem{fan2023lightgaussian}
Fan, Z., Wang, K., Wen, K., Zhu, Z., Xu, D., Wang, Z.: Lightgaussian: Unbounded 3d gaussian compression with 15x reduction and 200+ fps. arXiv preprint arXiv:2311.17245  (2023)

\bibitem{gao2019surfelwarp}
Gao, W., Tedrake, R.: Surfelwarp: Efficient non-volumetric single view dynamic reconstruction. In: Robotics: Science and Systems XIV (2019)

\bibitem{he2023h}
He, Z., Li, W., Zhang, T., Yuan, Y.: H 2 gm: A hierarchical hypergraph matching framework for brain landmark alignment. In: International Conference on Medical Image Computing and Computer-Assisted Intervention. pp. 548--558. Springer (2023)

\bibitem{hinton2015distilling}
Hinton, G., Vinyals, O., Dean, J.: Distilling the knowledge in a neural network. arXiv preprint arXiv:1503.02531  (2015)

\bibitem{huang2024endo}
Huang, Y., Cui, B., Bai, L., Guo, Z., Xu, M., Ren, H.: Endo-4dgs: Distilling depth ranking for endoscopic monocular scene reconstruction with 4d gaussian splatting. arXiv preprint arXiv:2401.16416  (2024)

\bibitem{kerbl20233d}
Kerbl, B., Kopanas, G., Leimk{\"u}hler, T., Drettakis, G.: 3d gaussian splatting for real-time radiance field rendering. ACM Transactions on Graphics  \textbf{42}(4) (2023)

\bibitem{lee2023compact}
Lee, J.C., Rho, D., Sun, X., Ko, J.H., Park, E.: Compact 3d gaussian representation for radiance field. arXiv preprint arXiv:2311.13681  (2023)

\bibitem{li2023steganerf}
Li, C., Feng, B.Y., Fan, Z., Pan, P., Wang, Z.: Steganerf: Embedding invisible information within neural radiance fields. In: Proceedings of the IEEE/CVF International Conference on Computer Vision. pp. 441--453 (2023)

\bibitem{li2022knowledge}
Li, C., Lin, M., Ding, Z., Lin, N., Zhuang, Y., Huang, Y., Ding, X., Cao, L.: Knowledge condensation distillation. In: European Conference on Computer Vision. pp. 19--35. Springer Nature Switzerland Cham (2022)

\bibitem{li2024endora}
Li, C., Liu, H., Liu, Y., Feng, B.Y., Li, W., Liu, X., Chen, Z., Shao, J., Yuan, Y.: Endora: Video generation models as endoscopy simulators. arXiv preprint arXiv:2403.11050  (2024)

\bibitem{li2024u}
Li, C., Liu, X., Li, W., Wang, C., Liu, H., Yuan, Y.: U-kan makes strong backbone for medical image segmentation and generation. arXiv preprint arXiv:2406.02918  (2024)

\bibitem{li2021unsupervised}
Li, C., Zhang, Y., Li, J., Huang, Y., Ding, X.: Unsupervised anomaly segmentation using image-semantic cycle translation. arXiv preprint arXiv:2103.09094  (2021)

\bibitem{li2021htd}
Li, W., Chen, Z., Li, B., Zhang, D., Yuan, Y.: Htd: Heterogeneous task decoupling for two-stage object detection. IEEE Transactions on Image Processing  \textbf{30},  9456--9469 (2021)

\bibitem{li2023novel}
Li, W., Guo, X., Yuan, Y.: Novel scenes \& classes: Towards adaptive open-set object detection. In: Proceedings of the IEEE/CVF International Conference on Computer Vision. pp. 15780--15790 (2023)

\bibitem{li2022scan}
Li, W., Liu, X., Yao, X., Yuan, Y.: Scan: Cross domain object detection with semantic conditioned adaptation. In: Proceedings of the AAAI Conference on Artificial Intelligence. vol.~36, pp. 1421--1428 (2022)

\bibitem{li2022sigma}
Li, W., Liu, X., Yuan, Y.: Sigma: Semantic-complete graph matching for domain adaptive object detection. In: Proceedings of the IEEE/CVF Conference on Computer Vision and Pattern Recognition. pp. 5291--5300 (2022)

\bibitem{li2020super}
Li, Y., Richter, F., Lu, J., Funk, E.K., Orosco, R.K., Zhu, J., Yip, M.C.: Super: A surgical perception framework for endoscopic tissue manipulation with surgical robotics. IEEE Robotics and Automation Letters  \textbf{5}(2),  2294--2301 (2020)

\bibitem{liang2021unsupervised}
Liang, Z., Rong, Y., Li, C., Zhang, Y., Huang, Y., Xu, T., Ding, X., Huang, J.: Unsupervised large-scale social network alignment via cross network embedding. In: Proceedings of the 30th ACM International Conference on Information \& Knowledge Management. pp. 1008--1017 (2021)

\bibitem{liu2024stereo}
Liu, X., Li, W., Yamaguchi, T., Geng, Z., Tanaka, T., Tsai, D.P., Chen, M.K.: Stereo vision meta-lens-assisted driving vision. ACS Photonics  (2024)

\bibitem{liu2021consolidated}
Liu, X., Guo, X., Liu, Y., Yuan, Y.: Consolidated domain adaptive detection and localization framework for cross-device colonoscopic images. Medical image analysis  \textbf{71},  102052 (2021)

\bibitem{liu2023efficientvit}
Liu, X., Peng, H., Zheng, N., Yang, Y., Hu, H., Yuan, Y.: Efficientvit: Memory efficient vision transformer with cascaded group attention. In: Proceedings of the IEEE/CVF Conference on Computer Vision and Pattern Recognition. pp. 14420--14430 (2023)

\bibitem{liu2024endogaussian}
Liu, Y., Li, C., Yang, C., Yuan, Y.: Endogaussian: Gaussian splatting for deformable surgical scene reconstruction. arXiv preprint arXiv:2401.12561  (2024)

\bibitem{liu2023grab}
Liu, Y., Li, W., Liu, J., Chen, H., Yuan, Y.: Grab-net: Graph-based boundary-aware network for medical point cloud segmentation. IEEE Transactions on Medical Imaging  (2023)

\bibitem{lu2021super}
Lu, J., Jayakumari, A., Richter, F., Li, Y., Yip, M.C.: Super deep: A surgical perception framework for robotic tissue manipulation using deep learning for feature extraction. In: ICRA. pp. 4783--4789. IEEE (2021)

\bibitem{luo2022unsupervised}
Luo, H., Wang, C., Duan, X., Liu, H., Wang, P., Hu, Q., Jia, F.: Unsupervised learning of depth estimation from imperfect rectified stereo laparoscopic images. Computers in biology and medicine  \textbf{140},  105109 (2022)

\bibitem{mahmoud2017orbslam}
Mahmoud, N., Cirauqui, I., Hostettler, A., Doignon, C., Soler, L., Marescaux, J., Montiel, J.M.M.: Orbslam-based endoscope tracking and 3d reconstruction. In: Computer-Assisted and Robotic Endoscopy: Third International Workshop, CARE 2016, Held in Conjunction with MICCAI 2016, Athens, Greece, October 17, 2016, Revised Selected Papers 3. pp. 72--83. Springer (2017)

\bibitem{mildenhall2021nerf}
Mildenhall, B., Srinivasan, P.P., Tancik, M., Barron, J.T., Ramamoorthi, R., Ng, R.: Nerf: Representing scenes as neural radiance fields for view synthesis. Communications of the ACM  \textbf{65}(1),  99--106 (2021)

\bibitem{pan2023learning}
Pan, P., Fan, Z., Feng, B.Y., Wang, P., Li, C., Wang, Z.: Learning to estimate 6dof pose from limited data: A few-shot, generalizable approach using rgb images. arXiv preprint arXiv:2306.07598  (2023)

\bibitem{peng2021low}
Peng, J., Sun, W., Li, H.C., Li, W., Meng, X., Ge, C., Du, Q.: Low-rank and sparse representation for hyperspectral image processing: A review. IEEE Geoscience and Remote Sensing Magazine  \textbf{10}(1),  10--43 (2021)

\bibitem{penza2017envisors}
Penza, V., De~Momi, E., Enayati, N., Chupin, T., Ortiz, J., Mattos, L.S.: Envisors: Enhanced vision system for robotic surgery. a user-defined safety volume tracking to minimize the risk of intraoperative bleeding. Frontiers in Robotics and AI  \textbf{4}, ~15 (2017)

\bibitem{rodriguez2022tracking}
Rodr{\'\i}guez, J.J.G., Montiel, J.M., Tard{\'o}s, J.D.: Tracking monocular camera pose and deformation for slam inside the human body. In: 2022 IEEE/RSJ International Conference on Intelligent Robots and Systems (IROS). pp. 5278--5285. IEEE (2022)

\bibitem{sherif2023remote}
Sherif, Y.A., Adam, M.A., Imana, A., Erdene, S., Davis, R.W.: Remote robotic surgery and virtual education platforms: How advanced surgical technologies can increase access to surgical care in resource-limited settings. In: Seminars in Plastic Surgery. Thieme Medical Publishers, Inc. (2023)

\bibitem{song2017dynamic}
Song, J., Wang, J., Zhao, L., Huang, S., Dissanayake, G.: Dynamic reconstruction of deformable soft-tissue with stereo scope in minimal invasive surgery. IEEE Robotics and Automation Letters  \textbf{3}(1),  155--162 (2017)

\bibitem{sun2022few}
Sun, L., Li, C., Ding, X., Huang, Y., Chen, Z., Wang, G., Yu, Y., Paisley, J.: Few-shot medical image segmentation using a global correlation network with discriminative embedding. Computers in biology and medicine  \textbf{140},  105067 (2022)

\bibitem{tang2018augmented}
Tang, R., Ma, L.F., Rong, Z.X., Li, M.D., Zeng, J.P., Wang, X.D., Liao, H.E., Dong, J.H.: Augmented reality technology for preoperative planning and intraoperative navigation during hepatobiliary surgery: a review of current methods. Hepatobiliary \& Pancreatic Diseases International  \textbf{17}(2),  101--112 (2018)

\bibitem{wang2022neural}
Wang, Y., Long, Y., Fan, S.H., Dou, Q.: Neural rendering for stereo 3d reconstruction of deformable tissues in robotic surgery. In: International Conference on Medical Image Computing and Computer-Assisted Intervention. pp. 431--441. Springer (2022)

\bibitem{wu20234dgaussians}
Wu, G., Yi, T., Fang, J., Xie, L., Zhang, X., Wei, W., Liu, W., Tian, Q., Xinggang, W.: 4d gaussian splatting for real-time dynamic scene rendering. arXiv preprint arXiv:2310.08528  (2023)

\bibitem{xu2022afsc}
Xu, H., Zhang, Y., Sun, L., Li, C., Huang, Y., Ding, X.: Afsc: Adaptive fourier space compression for anomaly detection. arXiv preprint arXiv:2204.07963  (2022)

\bibitem{yang2023neural}
Yang, C., Wang, K., Wang, Y., Yang, X., Shen, W.: Neural lerplane representations for fast 4d reconstruction of deformable tissues. arXiv preprint arXiv:2305.19906  (2023)

\bibitem{yang2023mrm}
Yang, Q., Li, W., Li, B., Yuan, Y.: Mrm: Masked relation modeling for medical image pre-training with genetics. In: Proceedings of the IEEE/CVF International Conference on Computer Vision. pp. 21452--21462 (2023)

\bibitem{yifan2019differentiable}
Yifan, W., Serena, F., Wu, S., {\"O}ztireli, C., Sorkine-Hornung, O.: Differentiable surface splatting for point-based geometry processing. ACM Transactions on Graphics (TOG)  \textbf{38}(6),  1--14 (2019)

\bibitem{yu2019dense}
Yu, W., Chen, H., Wang, L.: Dense attentional network for pancreas segmentation in abdominal ct scans. In: Proceedings of the 2nd International Conference on Artificial Intelligence and Pattern Recognition. pp. 83--87 (2019)

\bibitem{yu2022tnn}
Yu, W., Zheng, H., Gu, Y., Xie, F., Yang, J., Sun, J., Yang, G.Z.: Tnn: Tree neural network for airway anatomical labeling. IEEE Transactions on Medical Imaging  \textbf{42}(1),  103--118 (2022)

\bibitem{yu2022break}
Yu, W., Zheng, H., Zhang, M., Zhang, H., Sun, J., Yang, J.: Break: Bronchi reconstruction by geodesic transformation and skeleton embedding. In: 2022 IEEE 19th International Symposium on Biomedical Imaging (ISBI). pp.~1--5. IEEE (2022)

\bibitem{zha2023endosurf}
Zha, R., Cheng, X., Li, H., Harandi, M., Ge, Z.: Endosurf: Neural surface reconstruction of deformable tissues with stereo endoscope videos. In: International Conference on Medical Image Computing and Computer-Assisted Intervention. pp. 13--23. Springer (2023)

\bibitem{zhang2021generator}
Zhang, Y., Li, C., Lin, X., Sun, L., Zhuang, Y., Huang, Y., Ding, X., Liu, X., Yu, Y.: Generator versus segmentor: Pseudo-healthy synthesis. In: Medical Image Computing and Computer Assisted Intervention--MICCAI 2021: 24th International Conference, Strasbourg, France, September 27--October 1, 2021, Proceedings, Part VI 24. pp. 150--160. Springer International Publishing (2021)

\bibitem{zhou2021emdq}
Zhou, H., Jayender, J.: Emdq-slam: Real-time high-resolution reconstruction of soft tissue surface from stereo laparoscopy videos. In: MICCAI. pp. 331--340. Springer (2021)

\bibitem{zhuendogs}
Zhu, L., Wang, Z., Cui, J., Jin, Z., Lin, G., Yu, L.: Endogs: Deformable endoscopic tissues reconstruction with gaussian splatting

\end{thebibliography}

\end{document}